\documentclass{llncs}
\usepackage[utf8]{inputenc}
\usepackage{multirow}
\usepackage{booktabs}
\usepackage{rotating}
\usepackage{textcomp}
\usepackage[usenames,dvipsnames]{xcolor}

\newcommand{\textapprox}{\raisebox{0.5ex}{\texttildelow}}

\title{Deep Learning for Isotropic Super-Resolution from Non-Isotropic 3D Electron Microscopy}
\author{Larissa Heinrich, John A. Bogovic, Stephan Saalfeld}
\institute{HHMI Janelia Research Campus, Ashburn, USA}

\begin{document}
\maketitle
\begin{abstract}
The most sophisticated existing methods to generate 3D isotropic super-resolution (SR) from non-isotropic electron microscopy (EM) are based on learned dictionaries.  Unfortunately, none of the existing methods generate practically satisfying results. For 2D natural images, recently developed super-resolution methods that use deep learning have been shown to significantly outperform the previous state of the art.

We have adapted one of the most successful architectures (FSRCNN) for 3D super-resolution, and compared its performance to a 3D U-Net architecture that has not been used previously to generate super-resolution.

We trained both architectures on artificially downscaled isotropic ground truth from focused ion beam milling scanning EM (FIB-SEM) and tested the performance for various hyperparameter settings.

Our results indicate that both architectures can successfully generate 3D isotropic super-resolution from non-isotropic EM, with the U-Net performing consistently better. We propose several promising directions for practical application.
\end{abstract}

\section{Introduction}
Recent studies \cite{PlazaAl2014} argue that an isotropic resolution of less than 15\,nm per voxel is necessary to reconstruct the dense synaptic connectivity of entire animal nervous systems.  Today, two modes of 3D electron microscopy are available to generate volumetric image data at this resolution: (1) serial section EM tomography and (2) scanning EM in combination with focused ion beam milling (FIB-SEM).  Both modalities are comparably slow.  Imaging the entire central nervous system of a model organism as small as the fruit fly \emph{Drosophila melanogaster} takes many years on a single microscope \cite{hayworth2015}. Other EM methods such as serial section Transmission EM (ssTEM) or serial block-face scanning EM in combination with an automatic ultra-microtome and parallel acquisition \cite{mikula2016} are significantly faster but fail at generating the desired axial resolution.

To overcome this limitation, methods have been proposed to interpolate the missing
resolution from prior knowledge about the tissue, possibly in combination with sparse
tomography \cite{veeraraghavan2010}. However, even the most sophisticated proposals
\cite{glasner2011,hu2012}  are based on learning a discriminative, over-complete
dictionary using methods that are tuned for applications such as de-noising or
visually pleasing SR, and have not yet yielded practical improvements,
possibly due to the limited fields-of-view or generalizability.

Recently, several methods have been proposed that use deep learning for SR of 2D natural images \cite{dong2014,Ledig2016}.  Generally, these methods significantly outperform those based on learned dictionaries.  We adapted two successful convolutional neural network architectures (FSRCNN~\cite{dong2016FSRCNN} and U-Net \cite{ronneberger2015}) for image SR in the context of 3D electron microscopy.  Using high-resolution isotropic FIB-SEM data, we successfully trained them to predict high-resolution isotropic 3D images from non-isotropic input.  We compared the performance of these architectures with a set of different hyperparameters in terms of reconstruction accuracy and runtime.

\section{Related Work}
Dong et al.~\cite{dong2014} demonstrated that a simple convolutional
neural network architecture, mimicking a sparse-coding approach, could outperform those same methods
that were state-of-the-art at the time. Since then a number of deep-learning approaches were successful for SR of 2D natural images.  However, all deep-learning SR methods (that we are aware of) seek to increase the resolution
of both dimensions by the same factor. Methods for SR that take advantage of self-similarity
have seen great success and do not require high-resolution examples from
which to learn~\cite{Glasner2009}. 

In the context of electron microscopy (EM), Veerarghavan et
al.~\cite{veeraraghavan2010} and Hu et al.~\cite{hu2012} developed
methods for estimating a high-resolution segmentation from
lower-resolution EM imagery using over-complete dictionaries. These
methods leverage tomographic views of the tissue to be super-resolved,
which may not always be available.

The U-Net convolutional neural network architecture has had great success
in several pixel-wise prediction and segmentation tasks in biomedical
imaging~\cite{ronneberger2015}.  The key element of the U-Net are skip connections between a contracting path and an expanding path, thus providing multiple paths from input to output that incorporate features at different levels of resolution.

\section{Architecture Design}
\subsection{3D Anisotropic FSRCNN}
\begin{figure}
\centering
\includegraphics[width=0.95\textwidth]{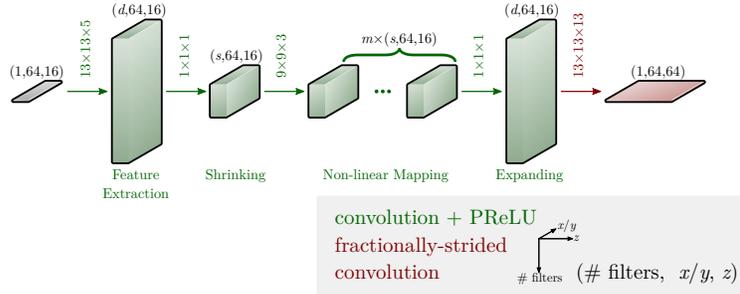}
\caption{Architecture of our 3D-FSRCNN with general hyperparameters $d$,
$s$, and $m$.}
\label{fig:fsrcnn}
\end{figure}
The FSRCNN \cite{dong2016FSRCNN} was designed for upscaling of 2D natural images. We adapted the architecture to the anisotropic 3D case (3D-FSRCNN) as shown in Figure~\ref{fig:fsrcnn}. 
The feature extraction layer in the FSRCNN is a single convolutional layer with a kernel size of $5\times5$, operating on the low resolution image. In the original SRCNN \cite{dong2014} the kernel size was $9\times9$ on the high resolution image for a scaling factor of 3. We thus chose to use a 3D convolution with a kernel size of $13\times13\times5$ for a scaling factor of 4 along the $z$-axis. For the kernel sizes of the non-linear mapping layers we used $9\times9\times3$ instead of $3\times3$ with the same reasoning. To be consistent with the feature extraction layer we used $13\times13\times13$ for the final fractionally-strided convolution (commonly referred to as deconvolution) layer, as opposed to $9\times9$ in the FSRCNN. The remaining specifications are consistent with the original FSRCNN.

Dong et al.~further identified three sensitive hyperparameters for their architecture: the number of filters in the feature extraction and expanding layer $d$, the number of filters in the shrinking and non-linear mapping layers $s$ and the number of non-linear mapping layers $m$. We investigated the same hyperparameters but increased the number of filters as ours need to represent 3D structures.

\subsection{3D Super-Resolution U-Net}
\begin{figure}[t]
\centering
\includegraphics[width=0.95\textwidth]{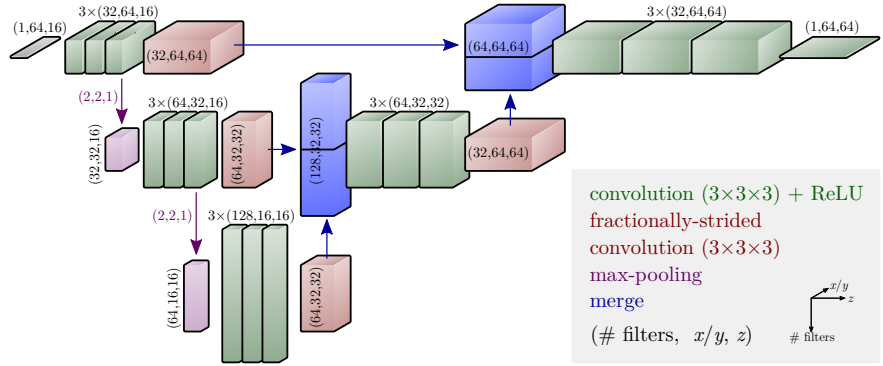}
\caption{Architecture of our 3D-SRU-Net with hyperparameters $w=32$, $h=3$, and
$d=3$.}
\label{fig:unet}
\end{figure}

The design of the U-Net architecture was motivated by the objective to
combine high localization accuracy and a large
field of view \cite{ronneberger2015}. As both of these are also crucial
factors for SR, we adapted the architecture for this purpose as shown in
Figure~\ref{fig:unet} (3D-SRU-Net).

On each level of the network in the contracting path, we introduced an additional fractionally-strided convolution layer such that the output of that layer becomes isotropic. Furthermore, we only downscaled along the lateral dimensions as long as the resolution in the axial dimension is lower. Instead of up-convolutions as in \cite{ronneberger2015} we used fractionally-strided convolutions as experiments showed that this reduced run-time without impairing performance. As in the original U-Net we doubled the number of features in each level and used the ReLU nonlinearity.  The free hyperparameters of this architecture are the number of levels $h$ (height), the initial number of filters $w$ (width), and the number of convolutional layers per level $d$ (depth).

\section{Experiments}
\subsection{Implementation Details} 

All experiments were carried out on a distortion free FIB-SEM dataset \cite{HanslovskyAl2016}. The FIB-SEM volume was downscaled from its native resolution of $8\times8\times2$\,nm to an isotropic resolution of $16\times16\times16$\,nm to reduce the influence of noise. The resulting volume had a size of $1250\times2000\times256$\,px and was divided into a training (70\%), a validation (15\%), and a test set (15\%) such that each set contained a variety of different textures.  We then simulated non-isotropic ssTEM data by downscaling the FIB-SEM volume on the fly by a factor of 4 along the $z$-axis using an average pooling layer as the first level in all networks (not shown in figures for brevity).

The networks were implemented in Python using \emph{keras} \cite{chollet2015} with the \emph{Tensorflow} \cite{tensorflow2016} backend. During training, random samples of size $64\times64\times64$\,px were drawn from the training data and downscaled to $64\times64\times16$\,px.  In order to speed up processing, all experiments used zero-padding such that the patch size was maintained.  At prediction time, small borders of the outputs were cut away to reduce the impact of border effects.

We used the Adam optimizer \cite{kingma2014adam} with a step-wise, square-root learning rate schedule. The learning rate for both network types is initialized to $\alpha_{\mathrm{init}}=10^{-4}$ with the 3D-FSRCNNs utilizing a faster decay. 

For the remaining optimizer parameters we followed the recommendations in \cite{kingma2014adam}. The batch size was 6 for all experiments, which is the largest value possible given our hardware (Nvidia Quadro M6000) for the network with the largest memory consumption.

The networks were trained to minimize the mean squared error (MSE), which is equivalent to maximizing the peak signal-to-noise ratio (PSNR). To emphasize the performance in `difficult' areas, we report a cubic-weighted PSNR (wPSNR) computed via a cubic-weighted MSE (wMSE), where the pixel-wise loss is weighted with the error resulting from cubic upsampling. The weighting is offset by a factor of $0.5$ to avoid values of $0$,
\begin{equation}
\mathrm{wMSE}= \frac{1}{XYZ} \sum_{x,y,z}^{X,Y,Z} \left (I^{\mathrm{HR}}_{x,y,z} - I^{\mathrm{pred}}_{x,y,z} \right)^2  \left(\frac{1}{2}+\frac{\left(I^{\mathrm{HR}}_{x,y,z} - I^{\mathrm{cub}}_{x,y,z}\right)^2}{2 \max\limits_{x,y,z}\left( \left(I^{\mathrm{HR}}_{x,y,z} - I^{\mathrm{cub}}_{x,y,z}\right)^2\right)}\right)
\end{equation}

\begin{tabular}{llll}
with$\quad$&$I^{\mathrm{HR}}$&: &high resolution image volume,\\
&$I^{\mathrm{pred}}$&: &prediction of image volume,\\
&$I^{\mathrm{cub}}$&: &cubic upsampling of image volume,\\
&$X,Y,Z$&: &dimensions of image volume.\\

\end{tabular}

\subsection{3D Anisotropic FSRCNN}

We tested twelve sets of hyperparameters ($m\in\{2,3,4\}$, $d\in\{240,280\}$, $s\in\{48,64\}$) for the 3D-FSRCNN architecture,
and measured their PSNR and weigh\-ted PSNR on the validation set (Table~\ref{tab:FSRCNN}) after \textapprox$290$k iterations.

Our experiments suggest that increasing the number of filters in the shrinking and mapping layers, $s$, has a slight positive effect on performance in most cases. This corresponds to larger dictionaries performing better in a sparse-coding framework. The number of filters in the feature extraction and expanding layer, $d$, had a very small effect on performance. Interestingly,
the depth parameter $m$ had a relatively small and inconclusive effect in contrast
to the conclusions drawn in the original FSRCNN paper.

\begin{table}[b]
\caption{PSNR, weighted PSNR, and number of parameters (nP) on the test set for 3D-FSRCNN models evaluated with the model snapshot providing the best PSNR on the validation set.
For comparison, cubic upsampling results in $\mathrm{PSNR}=33.22$ and $\mathrm{wPSNR}=35.94$.}
 
  \begin{center}
  \begin{tabular}{
    c
    c@{\hspace{1.0em}} 
    c@{\hspace{0.4em}}cc @{\hspace{0.8em}} 
    c@{\hspace{0.4em}}cc @{\hspace{0.8em}} 
    c@{\hspace{0.4em}}cc}
  \toprule
  &\multirow{2}{*}{Settings} & \multicolumn{3}{c}{$m=2$}& \multicolumn{3}{c}{$m=3$}&\multicolumn{3}{c}{$m=4$}\\
  &&PSNR&wPSNR&nP&PSNR&wPSNR&nP&PSNR&wPSNR&nP\\
  \midrule

  & $d(240)$, $s(48)$ &34.52 &37.28&  1.8m & 34.68 &37.45& 2.4m & 34.35 &37.12 & 2.9m \\
  & $d(280)$, $s(48)$  &34.42 &37.18& 2.0m &34.55 &37.32& 2.6m & 34.48 &37.25 & 3.1m \\
  & $d(240)$, $s(64)$ &34.53 &37.29& 2.8m & 34.35 &37.12 & 3.7m & \textbf{34.91} &\textbf{37.70} & 4.7m \\
  & $d(280)$, $s(64)$ &34.59 &37.36& 2.9m &34.77& 37.55& 3.9m & 34.71 &37.49 & 4.9m \\

  \bottomrule
  \end{tabular}
  \end{center}
\label{tab:FSRCNN}
\end{table}

\subsection{3D Super-Resolution U-Net}

In Table~\ref{tab:unet}, we report the PSNR and weighted PSNR for twelve sets of
hyperparameters ($h\in\{2,3,4\}$), $w \in \{32,64\}$, $d \in \{2,3\}$) of the 3D-SRU-Net architecture, all after \textapprox$290$k iterations. Increasing depth or width of the network both have a slight positive effect on the performance. Increasing the number of levels is particularly beneficial when going from $h=2$ to $h=3$. The step to $h=4$ only improves the performance for the smaller networks. The stagnating performance is likely connected to more overfitting in the largest networks. 
\begin{table}[t]
\caption{PSNR and weighted PSNR on the test set for 3D-SRU-Net models.  Twelve sets of hyperparameters were tested. }
\label{tab:unet}
\begin{center}
  \begin{tabular}{
    c@{\hspace{1.0em}}
    c@{\hspace{0.4em}}cc @{\hspace{0.8em}}
    c@{\hspace{0.4em}}cc @{\hspace{0.8em}}
    c@{\hspace{0.4em}}cc }
  \toprule
  \multirow{2}{*}{Settings} & \multicolumn{3}{c}{$h=2$}& \multicolumn{3}{c}{$h=3$}&\multicolumn{3}{c}{$h=4$}\\
  &PSNR&wPSNR&nP&PSNR&wPSNR&nP&PSNR&wPSNR&nP\\
  \midrule
  $d(2)$, $w(32)$ &35.68 &38.48& 2.3m & 36.09 &38.91 & 4.1m & 36.17 &39.00 & 8.9m \\
  $d(3)$, $w(32)$ &35.89 &38.71& 3.1m &36.21 &39.04 & 5.3m & 36.22&39.05 & 12.4m\\
  $d(2)$, $w(64)$&35.93 &38.74& 9.3m & 36.25 & 39.08& 16.2m & 36.22&39.04 & 35.7m \\
  $d(3)$, $w(64)$&36.21 &39.03& 12.2m & \textbf{36.27}& \textbf{39.09}& 21.3m & 36.25&39.08 & 49.6m \\
  \bottomrule
  \end{tabular}
\end{center}

\end{table}

\subsection{Comparison / Results}
Our hyperparameter comparison did not indicate a strong connection between model size and performance. Larger 3D-SRU-Net models tended to perform better which was not necessarily the case for the 3D-FSRCNNs.  For the two largest 3D-SRU-Nets, we did not observe a performance boost, presumably due to their increased tendency to overfit.  Particularly the 3D-FSRCNNs could potentially benefit from more tuning of optimization parameters.

All 3D-SRU-Nets clearly outperformed even the best 3D-FSRCNNs when trained for the
same number of iterations.  We conclude that this is not a model-size
issue alone.  We believe that the ``skip/merge-connections'' and
multi-scale representation make optimization easier (faster convergence)
for the 3D-SRU-Nets.  Both networks outperformed the sparse coding approach \cite{veeraraghavan2010} (Table~\ref{tab:sparsecoding}).\\
\begin{table}[b]
\caption{Evaluation of the dictionary learning algorithm by \cite{veeraraghavan2010} using different patch sizes $ps$, dictionary sizes $k$ and sparsity parameters $\lambda$.} 
\label{tab:sparsecoding}
\begin{center}
\begin{tabular}
{cccccccccc}
\toprule  
\multirow{3}{*}{Settings}&$ps$&8&12&12&12&12&8&16&16\\
&$k$&256&200&256&256&300&1024&256&256\\ &$\lambda$&0.05&0.05&0.03&0.05&0.05&0.07&0.05&0.09\\  
\midrule 
\multicolumn{2}{c}{PSNR}& 30.84&32.95&\textbf{33.92}&33.22&33.40&30.78&30.26&30.04\\
\multicolumn{2}{c}{wPSNR}&33.70&35.85&\textbf{36.81}&36.12&36.30&33.64&33.14&32.92\\
\multicolumn{2}{c}{nP}&131k&346k&442k&442k&518k&524k&1,049k&1,049k\\ 
\bottomrule 
\end{tabular}
\end{center}
\end{table}
When comparing wall clock time, the 3D-FSRCNN has an advantage due to its simple feed forward approach. Training the best-performing 3D-FSRCNN took 73.5h as opposed to 209h of training for the best 3D-SRU-Net. Nevertheless, training of the smallest 3D-SRU-Net was faster (67h) and showed superior performance.

\begin{figure}[t]
\centering
\includegraphics{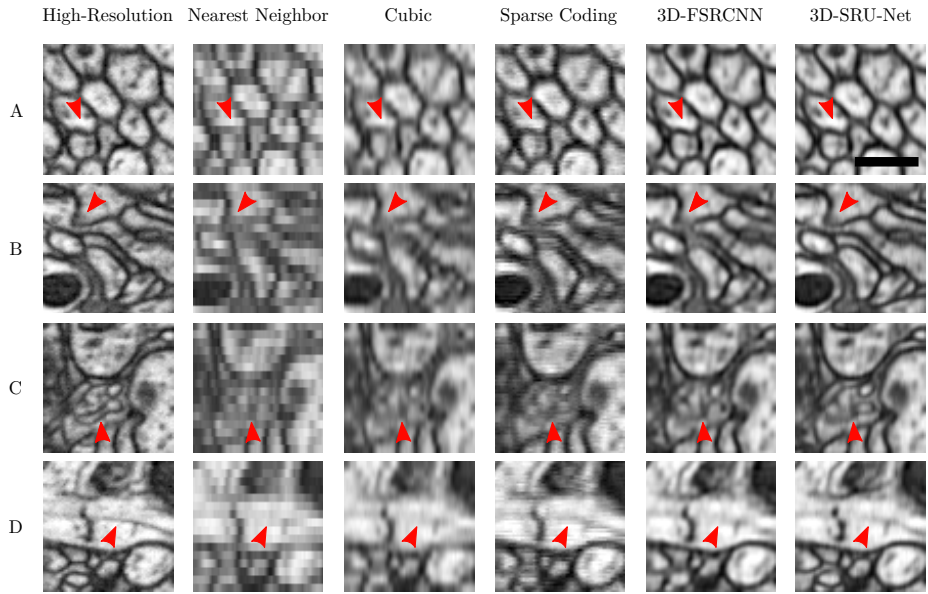}
\caption{Four examples of SR results for nearest neighbor and cubic interpolation,
the best-performing sparse coding, 3D-FSRCNN, and 3D-SRU-Net configurations.
Arrows indicate regions in which at least one SR result mis-interprets a
cell boundary or an ultrastructural feature. Scale bar 500\,nm.}
\label{fig:results}
\end{figure}

Figure~\ref{fig:results} shows exemplary image patches for both the best-performing 3D-FSRCNN ($m=4$, $d=240$, $s=64$) and the best-performing 3D-SRU-Net ($h=3$, $w=64$, $d=3$).
While the 3D-FSRCNN and the sparse coding approach produced reasonable results for common, simple textures (A) and consequently outperformed cubic upsampling, they failed to reconstruct more difficult textures (B and C). The 3D-SRU-Net architecture reliably reconstructed various structures (A and B), with only very rare exceptions (C). Sample D illustrates the effect on small ultrastructures (a microtubule) oriented along the lateral plane. The relatively strong smoothing is a common problem in SR but has been shown to be a consequence of the MSE loss. Including a perceptual loss function as in \cite{johnson2016} could potentially alleviate the issue.

\section{Conclusion}
In this work, we compared two deep convolutional neural network
architectures for 3D super-resolution.  The simple and fast 3D-FSRCNN
produced visually pleasing results in many cases.  The best performing
model (by PSNR) was a large 3D-SRU-Net architecture, perhaps unsurprisingly,
given its size, large field-of-view, and multi-scale
representation. While especially the 3D-FSRCNN could benefit from additional 
hyperparameter optimization, 
the 3D-SRU-Net design seems to overall be better suited for the task.
Both architectures outperform the sparse coding approach \cite{veeraraghavan2010}.

There is evidence that inferring high-resolution structure from
low-resolution imagery in EM connectomics is
possible~\cite{glasner2011}.  Our methodology could make segmentation
methods trained on isotropic FIB-SEM images applicable to anisotropic
ssTEM or block-face scanning EM volumes. The viability of the approach is an
interesting avenue for further research, though acquisition differences
between FIB-SEM and other EM methods may hinder this approach.
Self-similarity approaches to super-resolution~\cite{Glasner2009} have
been effective, but whether these ideas can generate effective 
training data for supervised CNNs is yet unknown.
We imagine that this approach could aid in automated segmentation
for connectomics as a kind of ``unsupervised pre-training.''
Manual annotation of training data for segmentation is incredibly
laborious\cite{PlazaAl2014}, and large CNNs typically require very many
labeled examples to learn from.  Yet, if a good representation can be 
learned to super-resolve, the same representation could be a good starting
point in training a network for segmentation.

\bibliographystyle{splncs03}
\bibliography{refs.bib}

\begin{thebibliography}{10}
\providecommand{\url}[1]{\texttt{#1}}
\providecommand{\urlprefix}{URL }

\bibitem{tensorflow2016}
Abadi, M., Barham, P., Chen, J., Chen, Z., et~al.: Tensorflow: A system for
  large-scale machine learning. In: 12th USENIX Symposium on Operating Systems
  Design and Implementation (OSDI 16). pp. 265--283 (2016)

\bibitem{chollet2015}
Chollet, F.: Keras. \url{https://github.com/fchollet/keras} (2015)

\bibitem{dong2014}
Dong, C., Loy, C.C., He, K., Tang, X.: Learning a deep convolutional network
  for image super-resolution. In: ECCV. pp. 184--199 (2014)

\bibitem{dong2016FSRCNN}
Dong, C., Loy, C.C., Tang, X.: Accelerating the super-resolution convolutional
  neural network. In: ECCV. pp. 391--407 (2016)

\bibitem{Glasner2009}
Glasner, D., Bagon, S., Irani, M.: {Super-resolution from a single image}. In:
  ICCV. pp. 349--356 (2009)

\bibitem{glasner2011}
Glasner, D., Hu, T., Nunez-Iglesias, J., Scheffer, L., et~al.: High resolution
  segmentation of neuronal tissues from low depth-resolution em imagery. In:
  International Workshop on Energy Minimization Methods in CVPR. pp. 261--272
  (2011)

\bibitem{HanslovskyAl2016}
Hanslovsky, P., Bogovic, J.A., Saalfeld, S.: Image-based correction of
  continuous and discontinuous non-planar axial distortion in serial section
  microscopy. Bioinformatics (btw794) (2016)

\bibitem{hayworth2015}
Hayworth, K.J., Xu, C.S., Lu, Z., Knott, G.W., et~al.: Ultrastructurally smooth
  thick partitioning and volume stitching for large-scale connectomics. Nature
  methods  12(4),  319--322 (2015)

\bibitem{hu2012}
Hu, T., Nunez-Iglesias, J., Vitaladevuni, S., Scheffer, L., et~al.:
  Super-resolution using sparse representations over learned dictionaries:
  Reconstruction of brain structure using electron microscopy. arXiv preprint
  arXiv:1210.0564  (2012)

\bibitem{johnson2016}
Johnson, J., Alahi, A., Fei-Fei, L.: Perceptual losses for real-time style
  transfer and super-resolution. In: European Conference on Computer Vision.
  pp. 694--711. Springer (2016)

\bibitem{kingma2014adam}
Kingma, D., Ba, J.: Adam: A method for stochastic optimization. In: ICLR (2015)

\bibitem{Ledig2016}
Ledig, C., Theis, L., Huszar, F., Caballero, J., o: {Photo-Realistic Single
  Image Super-Resolution Using a Generative Adversarial Network}. In:
  arXiv:1609.04802 (2016)

\bibitem{mikula2016}
Mikula, S.: Progress towards mammalian whole-brain cellular connectomics.
  Frontiers in Neuroanatomy  10, ~62 (2016)

\bibitem{PlazaAl2014}
Plaza, S.M., Scheffer, L.K., Chklovskii, D.B.: Toward large-scale connectome
  reconstructions. Current Opinion in Neurobiology  25,  201--210 (2014)

\bibitem{ronneberger2015}
Ronneberger, O., Fischer, P., Brox, T.: U-net: Convolutional networks for
  biomedical image segmentation. In: MICCAI. pp. 234--241 (2015)

\bibitem{veeraraghavan2010}
Veeraraghavan, A., Genkin, A.V., Vitaladevuni, S., Scheffer, L., et~al.:
  Increasing depth resolution of electron microscopy of neural circuits using
  sparse tomographic reconstruction. In: CVPR. pp. 1767--1774 (2010)

\end{thebibliography}

\end{document}